\documentclass[journal]{IEEEtran}
\ifCLASSINFOpdf
\else
\fi
\usepackage{amssymb}
\usepackage{amsmath}
\usepackage{booktabs}
\usepackage{multirow}
\usepackage{graphicx}
\usepackage{subfigure}
\usepackage{amsfonts,amssymb}
\usepackage{amsmath}
\usepackage{float}
\usepackage{color}

\usepackage{fancyhdr}

\begin{document}
\title{CAE-AV: Improving Audio-Visual Learning via Cross-modal Interactive Enrichment}


% \author{Michael~Shell,~\IEEEmembership{Member,~IEEE,}
%         John~Doe,~\IEEEmembership{Fellow,~OSA,}
%         and~Jane~Doe,~\IEEEmembership{Life~Fellow,~IEEE}}% <-this % stops a space
\author{Yunzuo~Hu,
        Wen~Li,
        and Jing~Zhang\IEEEauthorrefmark{1}%
\thanks{ Yunzuo Hu, Wen Li and Jing Zhang are with the School of Information Science and Engineering, East China University of Science and Technology (ECUST), Shanghai 200237, P. R. China.}%
\thanks{\IEEEauthorrefmark{1} Corresponding Author: jingzhang@ecust.edu.cn}}%
\maketitle

\begin{abstract}
  Audio-visual learning suffers from modality misalignment caused by off-screen sources and background clutter, and current methods usually amplify irrelevant regions or moments, leading to unstable training and degraded representation quality. To address this challenge, we proposed a novel Caption-aligned and Agreement-guided Enhancement framework (CAE-AV) for audio-visual learning, which used two complementary modules: Cross-modal Agreement-guided Spatio-Temporal Enrichment (CASTE) and Caption-Aligned Saliency-guided Enrichment (CASE) to relieve audio-visual misalignment. CASTE dynamically balances spatial and temporal relations by evaluating frame-level audio-visual agreement, ensuring that key information is captured from both preceding and subsequent frames under misalignment. CASE injects cross-modal semantic guidance into selected spatio-temporal positions, leveraging high-level semantic cues to further alleviate misalignment. In addition, we design lightweight objectives, caption-to-modality InfoNCE, visual-audio consistency, and entropy regularization to guide token selection and strengthen cross-modal semantic alignment. With frozen backbones, CAE-AV achieves state-of-the-art performance on AVE, AVVP, AVS, and AVQA benchmarks, and qualitative analyses further validate its robustness against audio-visual misalignment.
\end{abstract}


% \fi
\IEEEpeerreviewmaketitle

\section{Introduction}
\label{sec:intro}

\IEEEPARstart{A}{udio-Visual} Learning (AVL) jointly models visible content with spatial locality and acoustic sources with temporal cues, providing a unified perception-reasoning framework for tasks such as Audio-Visual Event Localization (AVE)~\cite{he2024cace,zhou2025towards,sun2025listen}, 
Audio-Visual Video Parsing (AVVP)~\cite{zhou2024label,sardari2024coleaf,zhao2025multimodal}, 
Audio-Visual Segmentation (AVS)~\cite{hao2024improving,mao2025contrastive,guo2025audio}, 
and Audio-Visual Question Answering (AVQA)~\cite{lao2023coca,park2024learning,li2025patch}, which has demonstrated significant value in applications including robotics, wearable assistants, and multimedia retrieval~\cite{avvp,avs,musicavqa}.

Recent studies show that parameter-efficient adaptation and early cross-modal interaction on top of frozen pretrained backbones can achieve a superior accuracy-efficiency trade-off ~\cite{lavish,dgdct,avmoe}. For example, LAVISH ~\cite{lavish} introduced cross-modal adapters within frozen backbones, establishing early interactions in latent space through bottleneck designs or lightweight attention to align and fuse features.
Duan et. al ~\cite{dgdct} employed cross-modal prompts or guidance signals, enabling each modality to perceive the other's semantics and structure during encoding, thereby improving focus on key frames and salient regions.
Inspired by Mixtures of Experts (MoE)~\cite{moe,mustafa2022multimodal,akbari2023alternating}, AVMOE ~\cite{avmoe} applied routing-based MoE mechanisms that dynamically allocate samples between uni-modal and cross-modal experts based on their characteristics, enhancing the task relevance of the representations.
Although above methods improve accuracy, efficiency, and scalability across multi-task settings while preserving the stability of frozen backbones, however, most methods assume strict framewise alignment between audio and video and overlook the common reality of misalignment. 

\begin{figure}[!t]
  \centering
  \includegraphics[width=1.0\linewidth]{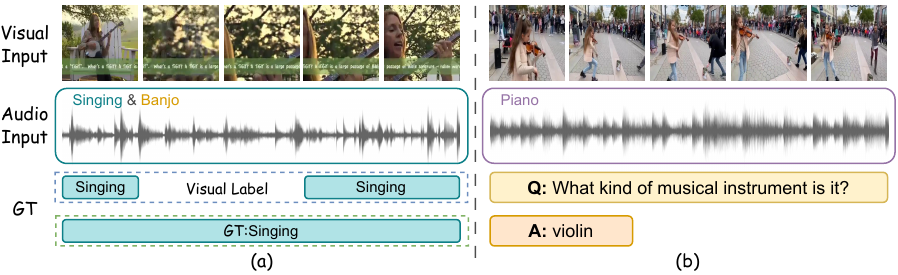}
  \caption{Two distinct audio-visual misalignment scenarios. (a) the singing continues while the camera cuts to a distant view, rendering the audio event temporarily invisible in the video. (b) a little girl is playing the violin, but the audio track is playing piano.}
  \label{fig:motivation}
\end{figure}

We systematically analyze a wide range of audio–visual misalignment cases. ~Fig.\ref{fig:motivation} highlights two representative examples. These two kinds of situations are common in multi-camera switching, B-roll, and post-production soundtracks, which frequently occur in open-world video.
Enforcing strict framewise alignment in these situations introduces incorrect moments or regions into the reasoning process, while overemphasizing cross-frame aggregation in alignment scenes can amplify irrelevant parts.
To address these issues, we try to operate at a fine per-frame granularity to avoid cross-frame interference in the alignment scenario. In the misalignment regime, the model must rely on stronger temporal modeling to aggregate evidence across frames, which can integrate the relative influence of spatial and temporal modeling and constrains where and how strongly cross-modal signals, enabling smooth transitions between the two scenarios.

Based on above analysis, we propose \textbf{C}onsistency-\textbf{A}ware \textbf{E}nrichment for \textbf{A}udio-\textbf{V}isual Learning (CAE-AV), which is built on frozen backbones and comprised two complementary enrichment pathways. We introduce a novel Cross-modal Agreement-guided Spatio-Temporal Enrichment (CASTE) module that leverages an Agreement Gate to estimate audio–visual consistency, generate spatial and temporal weights, recalibrate spatial features within each frame, and apply depthwise-separable 1D temporal convolutions to suppress noise propagation.
We also propose a Caption-Aligned Saliency-guided Enrichment (CASE) module that uses a frozen CLIP encoder~\cite{clip} to encode MLLM-generated captions and aligns feature via attention, followed by lightweight temporal consolidation for further more accurate spatiotemporal alignment.
In addition, the training process incorporates several loss terms, including InfoNCE based on subtitle-to-modal information, audiovisual consistency, and entropy regularization, which enforces semantic control and stabilize optimization while preserving the robustness of the frozen backbones. Our CAE-AV achieved state-of-the-art performance across a wide range of experiments on widely used public datasets on four downstream tasks including AVE, AVVP, AVS, and AVQA.

\section{Related Work}
\label{sec:related}

We briefly review two research directions that are most relevant to our approach: audio-visual learning and text-conditioned semantic priors.

% \subsection{Audio-Visual Learning}
\textbf{Audio-Visual Learning.}
Recent audio-visual learning studies focus on parameter-efficient fine-tuning, integrating cross-modal interactions into frozen pretrained backbones.
LAVISH~\cite{lavish} shows that a ViT pretrained only on images can be adapted to audio-visual tasks by inserting a small number of latent tokens and lightweight adapters layer-wise, which yields competitive performance across benchmarks with modest parameter growth. 
DG-SCT~\cite{dgdct} introduces cross-modal soft prompts that modulate spatial, channel, and temporal attention, enabling adaptive focus on informative cues without unfreezing the backbones.
AVMoE~\cite{avmoe} further adopts a routing-based mixture of uni-modal and cross-modal experts around frozen encoders, dynamically selecting experts per sample to enhance transferability and robustness with minimal trainable parameters.

All above methods are typically evaluated on standard audio-visual benchmarks. 
For \textbf{AVE}, the methods ~\cite{avsdn,cmran,cmbs,aveclip} all used post interaction strategies to better utilize visual and audio features encoded from pre trained models of specific modalities.
For \textbf{AVVP}, prior studies~\cite{mgn,yu2022mm,wu2021exploring} typically propose multi-scale hybrid networks, with feature aggregation conducted in a weakly supervised scenario.
For \textbf{AVS}, Mao et al.~\cite{ecmvae} employs a conditional multimodal VAE to disentangle shared and modality-specific representations while maximizing mutual information to improve alignment controllability. 
In \textbf {AVQA}, Li et al..~\cite{tspm} proposed a temporal-spatial perception model framework for locating time periods related to questions and enhancing spatial audio-visual associations, effectively demonstrating the reasoning process involved in answering questions.

Above parameter-efficient AVL mehtods usually uniformly process frames without explicit frame-wise agreement estimation. Under audio-visual misalignment, such early fusion often introduces noise into irrelevant regions, degrading representation quality.

\begin{figure*}
    \centering
    \includegraphics[width=0.7\linewidth]{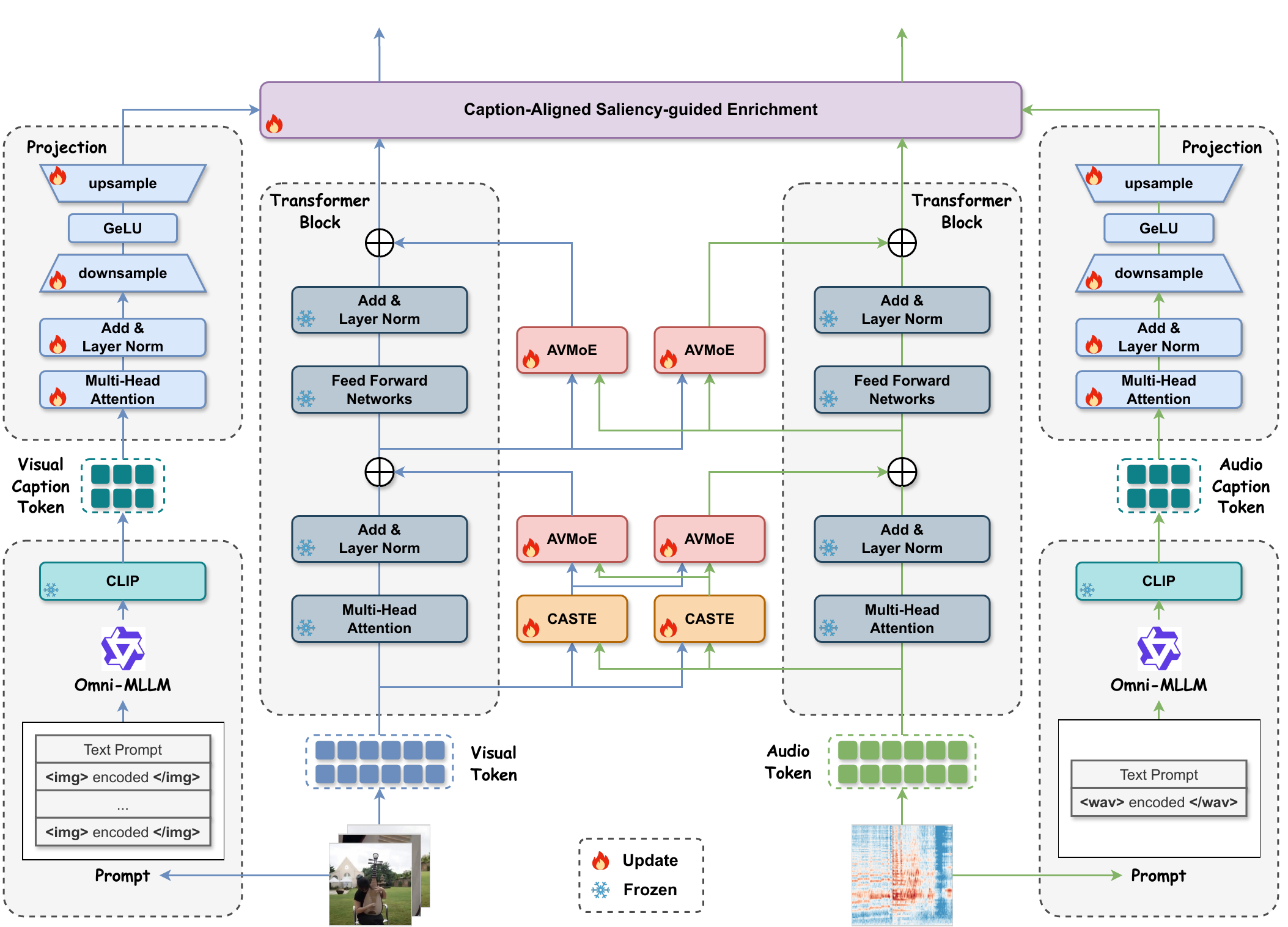}
    % \fbox{\rule{0pt}{2in} \rule{.9\linewidth}{0pt}}
    \caption{Method overview. CAE-AV injects CASTE and AVMoE into frozen pre-trained backbones. It constructs prompts for raw visual frames and audio to generate captions. Finally, the captions are projected into the feature space of the backbones to guide the inference of visual and audio features in CASE.}
    \label{fig:framework}
\end{figure*}

% \subsection{Text-conditioned Semantic Priors}
\textbf{Text-conditioned Semantic Priors.}
In multi-modal learning, leveraging semantic priors significantly enhances cross-modal alignment by providing high-level semantics, improving transferability and generalization across tasks~\cite{AudioCLIP,clap,mahmud2024t,guo2024open}.
CLIP~\cite{clip} aligns multi-modal through contrastive learning on large-scale image-text pairs, establishing text as a query interface. 
Recent studies~\cite{AudioCLIP, clap, mo2024audio} have incorporated audio into the CLIP framework, emphasizing contrastive pretraining for audio-text alignment, enabling cross-domain zero-shot generalization and produces versatile embeddings applicable. 
In the context of audio-visual tasks~\cite{mahmud2024t,guo2024open,liu2024valor}, text-based semantic priors now play a central role not only as an evaluation tool but also as a bridge for inference and retrieval. 
Above research demonstrates that injecting semantic priors can effectively enhance a model's cross-modal reasoning capabilities in multimodal scenarios dominated by audio-visual learning.

\textbf{Summary}. Current parameter-efficient AVL methods ~\cite{lavish, dgdct, avmoe} perform lightweight injection and dynamic selection on frozen backbone networks, but they cannot deal with audio-visual misalignment. Therefore, we propose a novel CAE-AV model to effectively relieve misalignment by two complementary paths: CASTE module performs consistency-guided spatiotemporal enrichment, and CASE module applies caption-aligned saliency injection. Based on these, CAE-AV delivers adaptive control over frozen backbones, while alignment and entropy-regularization losses stabilize training and inference.

\section{Method}
\label{sec:method}

%The overall framework of CAE-AV is illustrated in \ref{fig:framework}, which is built upon AVMoE~\cite{avmoe}. 
%AVMoE freezes large pretrained visual and audio backbones and inserts learnable mixture-of-experts adapters at every layer, including unimodal and cross-modal experts.  Then a lightweight router selects experts based on the spatiotemporal context and aggregates their outputs, preserving backbone stability and parameter efficiency while enabling cross-modal interaction and task adaptation.

To deal with the noise caused by audio-visual misalignment, we proposed CAE-AV, which improved AVMoE ~\cite{avmoe} as illustrated in Fig.\ref{fig:framework}. 
The backbones employed by CAE-AV are Swin Transformer~\cite{swint}, and HTS-AT~\cite{htsat}. 
Each layer consists of multi-head self-attention, a feed-forward network, and residual connections with layer normalization. 
To temper unreliable cross-modal signals, we place Cross-modal Agreement-guided Spatial Temporal Enrichment (CASTE) module before each Swin block is processed by the AVMoE adapters, which estimates audio-visual agreement and enriches audio or visual features that are more relevant at the moment by spatial and temporal emphasis. 
Following the backbones, we introduce Caption-Aligned Saliency-guided Enrichment (CASE), which leverages caption semantics as an anchor to align the modalities through lightweight, controllable injection, further eliminating audio-visual misalignment.

Next, we will introduce the input embeddings for the visual and audio, CASTE and CASE in detail.

\subsection{Audio-Visual Input Embeddings}
\label{sec:input_embeddings}

Following prior work~\cite{lavish,dgdct,avmoe}, we uniformly sample $T$ RGB frames from each video. Each frame is denoted by $I \in \mathbb{R}^{H \times W \times 3}$, where $H \times W$ is the spatial resolution. 
As in Swin Transformer~\cite{swint,liu2022swin}, each $I$ is partitioned into non-overlapping patches by a patch-splitting module with kernel size $(P_I, P_I)$. 
The patches are passed through a linear embedding layer to produce the initial visual tokens
$v^{(0)} \in \mathbb{R}^{T \times \frac{H}{P_I} \times \frac{W}{P_I} \times C_v}$,
where $C_v$ is the visual channel dimension.

For audio, around the timestamp of each video frame we extract a short waveform segment and compute a Mel-spectrogram $Ms \in \mathbb{R}^{L \times F}$, where $L$ is the number of time steps and $F$ is the number of frequency bins. 
Following HTS-AT~\cite{htsat}, the spectrogram is patchified by a Patch-Embed CNN with kernel size $(P_A, P_A)$ and then linearly projected to obtain the initial audio tokens
$a^{(0)} \in \mathbb{R}^{T \times \frac{L}{P_A} \times \frac{F}{P_A} \times C_a}$,
where $C_a$ is the audio channel dimension.

\subsection{Cross-modal Agreement-guided Spatial Temporal Enrichment}
\label{sec:caste}
In this subsection, we will describe how our proposed CASTE module estimates audiovisual consistency before determining the spatial and temporal focus of enrichment, thereby preventing cross-modal information injection from disrupting localization and temporal understanding. We explain the process primarily through the visual branch, with the audio branch following a symmetrical approach.
CASTE consists of four key components including the Agreement Gate, Spatial Enrichment, Temporal Enrichment, and Selective Injection. Next, we will introduce them in detail.

% \begin{figure}[ht]
%     \centering
%     % \fbox{\rule{0pt}{2in} \rule{0.9\linewidth}{0pt}}
%     \includegraphics[width=0.7\linewidth]{img/caste.pdf}

%     \caption{The CASTE architecture comprises Agreement Gate, Spatial Enrichment, Temporal Enrichment, and Selective Injection (L2 normalization \& Top-K).}
%     \label{fig:caste}
% \end{figure}

% \subsubsection{Agreement Gate}
\textbf{Agreement Gate.}
To obtain reliable cross-modal agreement before spatio and temporal enrichment, we compute framewise gates independently at layer $\ell$. 
Given visual and audio tokens $v^{(\ell)} \in \mathbb{R}^{T \times L_v \times C_v}$ and $a^{(\ell)} \in \mathbb{R}^{T \times L_a \times C_a}$, we first average tokens within each frame to form frame-level prototypes that stabilize global semantics and suppress intra-frame noise:
\begin{equation}
\bar v=\frac{1}{L_v}\sum_{i=1}^{L_v}v^{(\ell)}_{:,i,:} \quad
\bar a=\frac{1}{L_a}\sum_{j=1}^{L_a}a^{(\ell)}_{:,j,:}
\end{equation}
where $\bar v \in\mathbb{R}^{T\times C_v}$ and $\bar a \in\mathbb{R}^{T\times C_a}$. 
We then project both prototypes into a shared semantic space to remove modality-specific scale and dimensionality differences:
$
\tilde v=\bar v W_v + b_v$ and $\tilde a=\bar a W_a + b_a
$.

In the shared space, cross-modal agreement is measured by the cosine similarity, constrained to $[-1,1]$:
\begin{equation}
g=\left\langle \frac{\tilde v}{\|\tilde v\|_2},\ \frac{\tilde a}{\|\tilde a\|_2}\right\rangle
\end{equation}
with $\langle\cdot,\cdot\rangle$ denoting a framewise inner product. 
Because $g$ alone may be insufficient to disambiguate complex misalignment patterns, we construct a fusion vector that aggregates four complementary cues: 
$
\xi=[\ \tilde v,\ \tilde a,\ \tilde v-\tilde a,\ \tilde v\odot \tilde a\ ]\in\mathbb{R}^{T\times 4D}
$,
where $[\cdot,\cdot]$ denotes concatenation and $\odot$ the Hadamard product. 
The terms $\tilde v$ and $\tilde a$ preserve unimodal first-order evidence; $\tilde v-\tilde a$ encodes signed cross-modal discrepancies; $\tilde v\odot \tilde a$ emphasizes co-occurring, same-direction features that yield a more stable positive signal.
Feeding $\xi$ into a two-layer perceptron produces unnormalized preferences for spatial and temporal enrichment:
\begin{equation}
\lambda=\mathrm{MLP}(\xi)=W_2\sigma(W_1\xi+b_1)+b_2\in\mathbb{R}^{T\times 2},
\end{equation}
where $\sigma(\cdot)$ is ReLU. 

To inject a mild prior early in training, we bias $\lambda$ with $[g,-g]$, encouraging frames with higher agreement to favor spatial modeling and frames with lower agreement to favor temporal modeling:
\begin{equation}
[w_{sp},w_{tm}]=\mathrm{softmax}(\lambda+[g,-g]), w_{sp}+w_{tm}=1,
\end{equation}
where the softmax operates over the last dimension. The resulting weights $w_{sp}$ and $w_{tm}$ are broadcast over tokens within each frame and used to form a convex combination of the spatial- and temporal-enrichment outputs. 
This gating-with-bias mechanism markedly reduces erroneous injection under misalignment and allows the subsequent spatial and temporal modules to specialize, improving localization and temporal reasoning.

% \subsubsection{Spatial Enrichment}
\textbf{Spatial Enrichment.}
In audio-visual learning, trainable units added to a frozen visual backbone must remain small in parameter count and conservative in update magnitude. Otherwise the pretrained representation is easily disturbed and cross-modal noise is amplified. 
Guided by this principle, the spatial enrichment module operates within each frame using separable attention in MobileViT~\cite{mobilev2}. 
It performs token to global recalibration with linear complexity and very few learnable parameters, avoiding the quadratic pairwise token interactions and memory cost of standard multihead self attention.

Consider layer $\ell$ and for a given frame $t$, the visual tokens are $v^{(\ell)}_t \in \mathbb{R}^{L_v \times C_v}$. 
Let the learnable projections be
$W_i \in \mathbb{R}^{C_v \times 1}$,
$W_k \in \mathbb{R}^{C_v \times C_v}$,
$W_v \in \mathbb{R}^{C_v \times C_v}$,
$W_o \in \mathbb{R}^{C_v \times C_v}$,
and let $\sigma(\cdot)$ denote the logistic function. 
% We first compute an Input $\mathcal{I}$, Key $\mathcal{K}$, and Value $\mathcal{V}$ to summarize frame-level context:
To summarize frame-level context, we first compute an Input $\mathcal{I}=\sigma\left(v^{(\ell)}_t W_i\right)$, Key $\mathcal{K}=v^{(\ell)}_t W_k$, and Value $\mathcal{V}=v^{(\ell)}_t W_v$.
With $\mathcal{I}_p \in \mathbb{R}$ and $\mathcal{K}_p \in \mathbb{R}^{1\times C_v}$ denoting the input and key of token $p$, a global context vector is obtained by a gate-weighted sum and this context recalibrates the values to suppress diffuse noise and highlight salient regions:
\begin{equation}
\mathcal{C}=\sum_{p=1}^{L_v} \mathcal{I}_p \odot \mathcal{K}_p, \quad
% \end{equation}
% \begin{equation}
\hat{\mathcal{V}}  = \mathcal{V} \odot \mathcal{C}.
\end{equation}

To inject a small, reversible update alongside the frozen backbone, we apply layer normalization and a residual gate. 
Let $\gamma_{sp}$ be a learnable scalar and $LN(\cdot)$ be layer normalization, the spatial enrichment output is
\begin{equation}
    v_{sp}^{(\ell)}=\mathrm{LN}( [v^{(\ell)}_i+\gamma_{sp} \hat{\mathcal{V}}_i W_o]_{i \in [1,T]} ).
\end{equation}
Similarly, we perform spatial enrichment on the audio to obtain $a_{sp}^{(\ell)}$.

% \subsubsection{Temporal Enrichment}
\textbf{Temporal Enrichment.}
Temporal enrichment models short-range dynamics across frames with minimal trainable overhead while complementing the frozen backbone's temporal capacity. 
We treat the temporal trajectory of each spatial position as a sequence of length $T$. 
A depthwise one-dimensional convolution \cite{mobilev1} is applied only along time, which aggregates information in the local neighborhood $(t-1, t, t+1)$ and makes the temporal receptive field explicit.

Let $\mathcal{D}_k$ denote a depthwise 1D convolution of kernel size $k$ along time that smooths and captures variations per channel, and let $\mathcal{P}_k$ denote a pointwise 1D convolution that linearly mixes channels at each time step. 
The computation is
\begin{equation}
Z=Reshape(\mathcal{P}_1 ( \mathcal{D}_3(Reshape(v^{(\ell)})))) ,
\label{eq:conv}
\end{equation}
\begin{equation}
v_{tm}^{(\ell)}=\mathrm{LN}(v^{(\ell)}+\gamma_{tm}\,Z),
\end{equation}
where $Reshape$ reorders axes to expose the temporal dimension for convolution and then restores the original layout, and $\gamma_{tm}$ is a learnable scalar gate. 
Similarly, we perform temporal enrichment on the audio to obtain $a_{tm}^{(\ell)}$.

% Because the convolution operates strictly along time and aligns adjacent frames at the same spatial position, it naturally emphasizes onsets and offsets, abrupt changes, and short-range rhythms, while remaining robust to small misalignments and noise and keeping computation low.

% \subsubsection{Selective Injection}
\textbf{Selective Injection.}
After estimating the framewise weights, we form an interpretable convex combination within each frame:
$
\hat v^{(\ell)} = w_{sp}v_{sp}^{(\ell)} + w_{tm}v_{tm}^{(\ell)}.
$
To avoid uniform injection when information quality varies across locations, we define token saliency by energy and apply sparse selection. 
Let $\|\cdot\|_2$ denote the $\ell_2$ norm over channels. The saliency of token $i$ in frame $t$ is
$
s_{t,i}=\|\hat v^{(\ell)}_{t,i,:}\|_2^2,
$
and the Top-$K$ tokens per frame are chosen to form a binary mask $M\in\{0,1\}^{T\times L_v\times 1}$, where $K=\lceil \rho,L_v\rceil$ with $\rho=0.3$. 
The mask is broadcast along the channel dimension and applied to $\hat v^{(\ell)}$, so only the most discriminative positions receive enrichment. 
This improves localization and alignment while keeping parameter and gradient budgets under control:
\begin{equation}
v_{caste}^{(\ell)}=v^{(\ell)} + \gamma(\hat v^{(\ell)} \odot M),
\end{equation}
where $\gamma$ is a learnable gate.

After obtaining the injected features for both modalities, we pass them to AVMoE and write them back to their respective backbones to close the loop within the layer:
\begin{equation}
v_{caste}^{(\ell)} = \Omega_{v}^{(\ell)}(v^{(\ell)}, a^{(\ell)}),
a_{caste}^{(\ell)} = \Omega_{a}^{(\ell)}(a^{(\ell)}, v^{(\ell)}),
\end{equation}
\begin{equation}
v_{moe}^{(\ell)} = \Phi_{v}^{(\ell)}(v_{caste}^{(\ell)}, a_{caste}^{(\ell)}),
a_{moe}^{(\ell)} = \Phi_{a}^{(\ell)}(a_{caste}^{(\ell)}, v_{caste}^{(\ell)}),
\end{equation}
\begin{equation}
v^{(\ell)} = v^{(\ell)} + \mathrm{MHA}(v^{(\ell)}) + v_{moe}^{(\ell)},
\end{equation}
\begin{equation}
a^{(\ell)} = a^{(\ell)} + \mathrm{MHA}(a^{(\ell)}) + a_{moe}^{(\ell)},
\end{equation}
where $\Omega^{(\ell)}$ denotes the CASTE module of the $\ell$th layer, $\Phi^{(\ell)}$ denotes the AVMoE module of the $\ell$th layer, and $\mathrm{MHA}(\cdot)$ denotes the multi-head attention.
For brevity, the equations omit layer normalization and feed-forward networks.

\subsection{Caption Obtaining and Processing}
\label{sec:caption}

Captions from audio and video can effectively help us identify audio-visual misalignment, hence we propose to fully use captions to relieving this issue.
Here we will introduce how to obtain and process captions of visual and audio. As shown at the far left and right of Fig.\ref{fig:framework}, we first use the same multimodal large language model (MLLM) to transcribe original visual and audio data into text captions. 
These captions are then encoded into vector sequences via a frozen CLIP~\cite{clip} model.

We adopt an MLLM to generate captions rather than directly extracting features from the MLLM for several key reasons. First, captions serve as an explicit intermediate representation that offers greater flexibility and controllability. Once generated, captions can be post-processed, edited, and filtered, which helps stabilize and enhance the quality of textual features. In contrast, directly using the MLLM’s latent features tends to be more opaque, making them difficult to interpret and constrain. Second, this approach provides stronger interpretability. The generated captions are readable semantic evidence, facilitating analysis of the sources of model decisions and error attribution. Third, from a system perspective, although caption generation introduces an additional step, the CLIP text encoder is typically lightweight and efficient, whereas MLLMs are often much larger and computationally more complex. Directly extracting MLLM features could require extra cross-modal forward computations, intermediate-layer readouts, or feature alignment operations, resulting in substantially higher overall costs. Finally, to enable a more comparable experimental setup, we avoid using the MLLM’s latent features as the core representation, thereby reducing strong dependence on a particular large model’s capabilities and improving fairness and reproducibility to some extent.

Then a lightweight mapping layer projects these sequences into a space compatible with the backbone features, enabling their subsequent use as high-level semantic anchors for cross-modal alignment in CASE. 
To mitigate systemic biases arising from different generators and decoding strategies, we employ the same MLLM across both modalities while maintaining consistent prompt templates and decoding hyperparameters. 
This ensures text distributions on both ends remain symmetrical in style and length, thereby treating both branches equitably during training and evaluation.
Detailed procedures and prompts are provided in Supplementary Section 1.

Given a sequence of video frames ${\{I_t\}}_{t=1}^{T}$ and the corresponding audio wave $S$, we base64-encode each modality and concatenate it with a neutral prompt. 
Let $\Gamma(\cdot;\tau)$ denote the MLLM captioning operator with temperature $\tau$. 
The resulting captions are
\begin{equation}
\mathrm{Cap}_v=\Gamma(\pi_v({\{I_t\}}_{t=1}^{T});\tau),
\mathrm{Cap}_a=\Gamma(\pi_a({S});\tau),
\end{equation}
where $\pi_v(\cdot)$ and $\pi_a(\cdot)$ are modality-symmetric prompt templates designed to reduce prompt-induced bias. 
Here we use Qwen-2.5-omni~\cite{qwen} as the Omni-MLLM.

Then a frozen CLIP text encoder $E_{\text{clip}}$ then maps the captions to token sequences:
$
S_{\text{ori},v}=E_{\text{clip}}(\mathrm{Cap}_v),
$
and
$
S_{\text{ori},a}=E_{\text{clip}}(\mathrm{Cap}_a)
$.
Freezing CLIP keeps the semantic anchors stable and decoupled from downstream optimization, mitigating overfitting to the task data in the language space.

To align caption features with the channel dimensions of the visual and audio backbones, we employ a shared self-attention module followed by modality-specific bottleneck projection modules. 
A single set of multi-head self-attention parameters $\Theta_{\text{att}}$ denoises both caption sequences, with layer normalization for stability.
% First, perform lightweight denoising on both end captions using the same set of multi-head self-attention parameters $\Theta_{\text{att}}$, combined with layer normalization for stabilization. 
% This parameter sharing aims to process both texts within the same attention subspace, thereby reducing style drift and semantic inconsistencies caused by independent parameterization. 
\begin{equation}
S'_v=\mathrm{LN}(S_{\text{ori},v}+\mathrm{MHA}(S_{\text{ori},v};\Theta_{\text{att}})),
\end{equation}
\begin{equation}
S'_a=\mathrm{LN}(S_{\text{ori},a}+\mathrm{MHA}(S_{\text{ori},a};\Theta_{\text{att}})).
\end{equation}
Sharing $\Theta_{\text{att}}$ encourages both sides to be parsed within the same attention subspace, reducing style drift and semantic inconsistency that could arise from independent parameterization.
We then project the caption channel dimension $C_t$ to the backbone channel sizes $C_v$ and $C_a$ using linear-GELU-linear bottlenecks. 
% With $\phi(\cdot)$ denoting GELU and ${W_{\text{down}},W_{\text{up}}}$ denoting the projection weights:
% \begin{equation}
% \tilde S_v=\phi\big(S'_v W^{v}_{\text{down}}\big)W^{v}_{\text{up}},\qquad
% \tilde S_a=\phi\big(S'_a W^{a}_{\text{down}}\big)W^{a}_{\text{up}}.
% \end{equation}
Only the self-attention $\Theta_{\text{att}}$ is shared and the output projections are modality-specific, which preserves a unified semantic re-estimation stage while allowing flexible adaptation to each backbone's channel space.

Finally, to align with frame-level features, we repeat the caption tokens along time to length $T$ to get $v_{cap}$ and $a_{cap}$.
% \begin{equation}
% v_{cap}=\mathrm{Repeat}(\tilde S_v,T),
% a_{cap}=\mathrm{Repeat}(\tilde S_a,T),
% \end{equation}
These features serve as high-level semantic guidance for CASE in the subsequent alignment stage.

\subsection{Caption-Aligned Saliency-guided Enrichment}
\label{sec:case}

\begin{figure}[ht]
    \centering
    \includegraphics[width=0.7\linewidth]{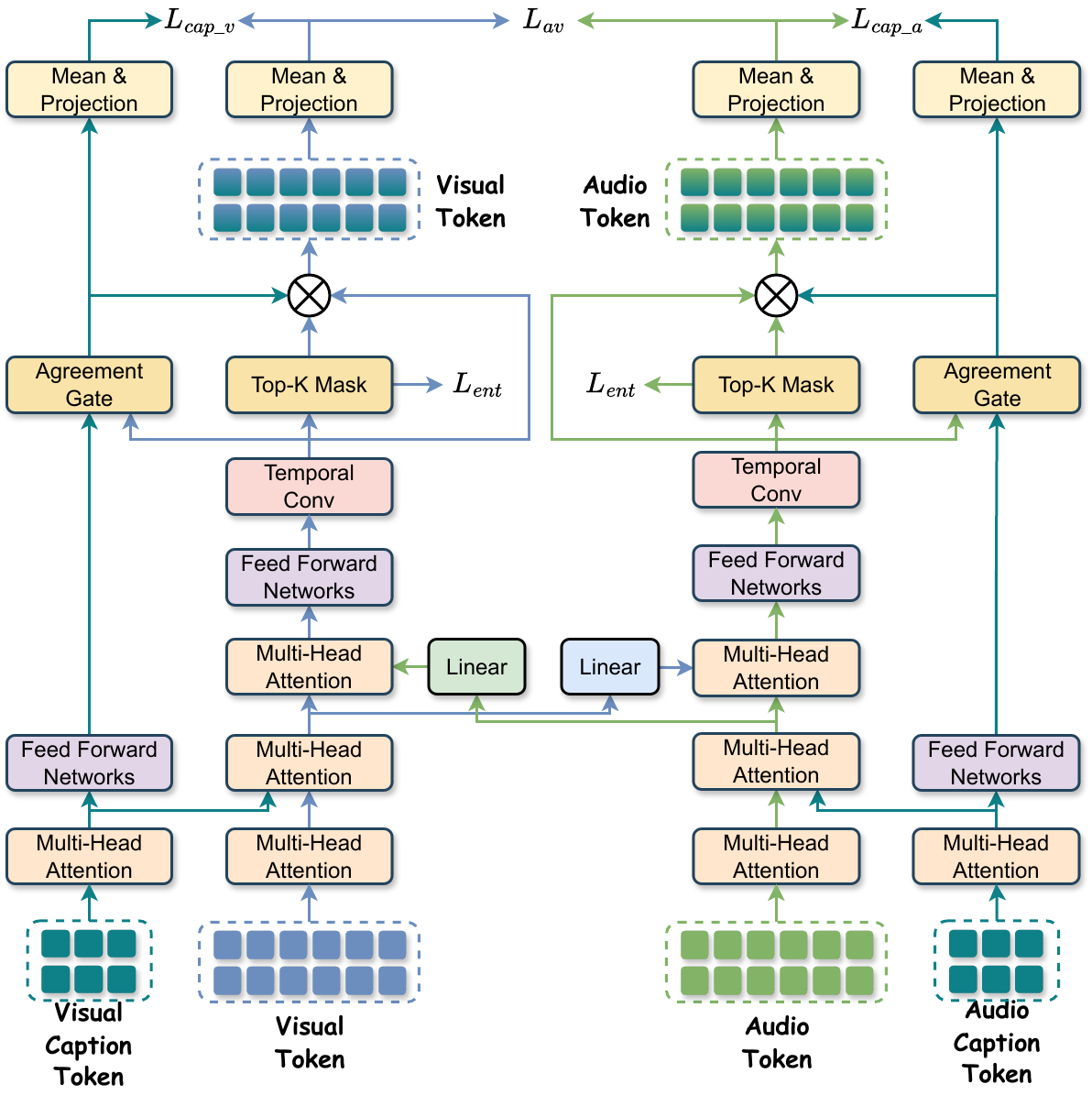}

    \caption{The structure of CASE and the calculation of loss.}
    \label{fig:case}
\end{figure}

To introduce stable high-level semantic anchors under frozen backbones and correct cross-modal attention drift, we propose a new Caption-Aligned Saliency-guided Enrichment (CASE) module following with the backbones, as shown in Fig.\ref{fig:case}. 
The $v^{(L)}$ and $a^{(L)}$ denote the frame-level video and audio tokens from the backbones, and $v_{\text{cap}}$ and $a_{\text{cap}}$ are the caption features projected to the backbone channel sizes, which is described in Section \ref{sec:caption}.

% Caption tokens act as semantic providers and backbone tokens act as queries. 
We first apply multi-head self-attention to both the caption and backbone tokens, followed by a cross-attention operation. The resulting features enriched with caption information are denoted as $\tilde{v}$ and $\tilde{a}$, respectively.
% \begin{equation}
% v_{\text{cap}} = \mathrm{MHA}(v_{\text{cap}}), \quad
% a_{\text{cap}} = \mathrm{MHA}(a_{\text{cap}})
% \end{equation}
% \begin{equation}
% v^{(L)} = \mathrm{MHA}(v^{(L)}), \quad
% a^{(L)} = \mathrm{MHA}(a^{(L)})
% \end{equation}
% \begin{equation}
% \tilde v=\mathrm{MHCA}\big(v^{(L)},v_{\text{cap}},v_{\text{cap}}\big)
% \end{equation}
% \begin{equation}
% \tilde a=\mathrm{MHCA}\big(a^{(L)},a_{\text{cap}},a_{\text{cap}}\big)
% \end{equation}
% where $\mathrm{MHCA}(Q,K,V)$ denotes multi-head cross-attention.
To propagate complementary cues across modalities, we project $\tilde{a}$ to the visual feature dimension and $\tilde{v}$ to the audio feature dimension using linear mappings $W^{v \leftarrow a}$ and $W^{a \leftarrow v}$, respectively. The features are then symmetrically refined symmetrically as follows:
\begin{equation}
\hat v=\mathrm{MHCA}\big(\tilde v,\ \tilde a W^{v\leftarrow a},\ \tilde a W^{v\leftarrow a}\big),
\end{equation}
\begin{equation}
\hat a=\mathrm{MHCA}\big(\tilde a,\ \tilde v W^{a\leftarrow v},\ \tilde v W^{a\leftarrow v}\big),
\end{equation}
where $\mathrm{MHCA}(Q,K,V)$ denotes multi-head cross-attention.
Depthwise separable temporal convolutions followed by pointwise mixing smooth short-range dynamics and suppress jitter, enabling us to obtain: 
$
\bar v=\mathcal{T}_v(\hat v)
$
and
$
\bar a=\mathcal{T}_a(\hat a)
$
with $\mathcal{T}_v$ and $\mathcal{T}_a$ defined as in Eq.\ref{eq:conv}.

Caption-feature agreement sets the injection strength. For each frame $t$, we mean-pool tokens, $\ell_2$-normalize, and compute cosine similarity as follows:
\begin{equation}
    r_v(t)= 
    \Big\langle
    \frac{\mathrm{mean}_i(\bar v_{t,i,:})}{\|\mathrm{mean}_i(\bar v_{t,i,:})\|^2},
    \frac{\mathrm{mean}_j(v_{\text{cap},t,j,:})}{\|\mathrm{mean}_j(v_{\text{cap},t,j,:})\|^2}
    \Big\rangle,
\end{equation}
and define frame gates $g_v(t)=\sigma(r_v(t))$ and $g_a(t)=\sigma(r_a(t))$ where $\sigma$ denotes Sigmoid function, broadcast along tokens. 
Token saliency selects informative positions by computing scores $s_{t,i}^{(v)}=\|\bar v_{t,i,:}\|_2^2$ and $s_{t,i}^{(a)}=\|\bar a_{t,i,:}\|_2^2$, from which we take the Top-$K$ per frame to form binary masks $M_v$ and $M_a$.
The selective injection is
\begin{equation}
v_{\text{case}}=v+\gamma_v\big(g_v\odot \mathrm{LN}(\bar v)\odot M_v\big),
\end{equation}
\begin{equation}
a_{\text{case}}=a+\gamma_a\big(g_a\odot \mathrm{LN}(\bar a)\odot M_a\big),
\end{equation}
where $\gamma_v$ and $\gamma_a$ are learnable residual gates. 
The CASE turns high-level caption concepts into differentiable constraints at both frame and token levels, amplifying cross-modal signals only where evidence is sufficient and consistent, which improves localization and temporal reasoning with minimal disturbance to the frozen backbones.

\textbf{Loss Functions.} 
% To further stabilize semantic alignment, we add a lightweight set of contrastive and consistency losses on top of CASE. Let the projection heads be
% $P_v:\mathbb{R}^{C_v}\to\mathbb{R}^{d}$,
% $P_a:\mathbb{R}^{C_a}\to\mathbb{R}^{d}$,
% $P_{vc}:\mathbb{R}^{C_v}\to\mathbb{R}^{d}$,
% $P_{ac}:\mathbb{R}^{C_a}\to\mathbb{R}^{d}$,
% and let $\tau>0$ be the temperature. We average tokens within each frame and project:
% \begin{equation}
% z_{v,t}=P_v(\mathrm{mean}(v_{case})),
% z_{a,t}=P_a(\mathrm{mean}(a_{cases}))
% \end{equation}
% \begin{equation}
% z_{vc,t}=P_{vc}(\mathrm{mean}(v_{cap})),
% z_{ac,t}=P_{ac}(\mathrm{mean}(a_{cap})).
% \end{equation}
To stabilize alignment, we add lightweight contrastive and consistency losses to CASE. Using projection heads $P_v, P_a, P_{vc}, P_{ac}$, we compute frame-averaged projections: $z_{v,t}=P_v(\mathrm{mean}(v_{case}))$, $z_{a,t}=P_a(\mathrm{mean}(a_{case}))$, $z_{vc,t}=P_{vc}(\mathrm{mean}(v_{cap}))$, $z_{ac,t}=P_{ac}(\mathrm{mean}(a_{cap}))$.

To prevent frames from the same video acting as easy negatives, we remove within-video negatives from InfoNCE. 
Let the positive pair for frame $t$ be $(z_{v,t}, z_{vc,t})$. Let $\mathcal{N}(t)$ be the set of negatives that excludes frames from the same video. 
With $\hat z = z/\|z\|_2$ and $N$ the number of valid frames in the batch, the visual-caption alignment loss is
\begin{equation}
\mathcal{L}_{\text{cap}}^{v}
=-\frac{1}{N}\sum_{t}\log\frac{\exp\big(\langle \hat z_{v,t},\hat z_{vc,t}\rangle/\tau\big)}
{\sum_{t'\in\mathcal{N}(t)\cup{t}}\exp\big(\langle \hat z_{v,t},\hat z_{vc,t'}\rangle/\tau\big)},
\end{equation}
where  $\tau>0$ is the temperature.
The audio-caption loss $\mathcal{L}_{\text{cap}}^{a}$ is defined analogously with $(z_a, z_{ac})$.

To enforce cross-modal semantic alignment within the same frame, we minimize the cosine distance between the modality-specific representations as follows:
\begin{equation}
\mathcal{L}_{\text{va}}=1-\frac{1}{N}\sum_{t}\langle \hat z_{v,t},\hat z_{a,t}\rangle.
\end{equation}

To encourage sharper Top-$K$ selections, we regularize the soft selection probabilities using a negative entropy term. Specifically, let $p_{v,t}$ and $p_{a,t}$ denote the selection distributions for the visual and audio modalities at frame $t$. Given the entropy definition $H(p) = -\sum_i p_i \log p_i$, the entropy regularization loss is formulated as:
\begin{equation}
\mathcal{L}_{\text{ent}}=-\frac{1}{N}\sum_{t}\big(H(p_{v,t})+H(p_{a,t})\big).
\end{equation}

The overall objective for CASE is defined as follows:
\begin{equation}
\label{eq:loss}
\mathcal{L}_{\text{CASE}}=
\alpha\big(\mathcal{L}_{\text{cap}}^{v}+\mathcal{L}_{\text{cap}}^{a}\big)
+\beta\mathcal{L}_{\text{va}}
+\gamma\mathcal{L}_{\text{ent}},
\end{equation}
where $\alpha,\beta,\gamma$ are scalar weights and we use staged weight scheduling to avoid over-constraining the frozen backbones early in training.

\section{Experiments}
\label{sec:experiments}

We evaluate CAE-AV on four representative downstream tasks, including audio-visual question answering (AVQA), audio-visual segmentation (AVS), audio-visual event localization (AVE), and audio-visual video parsing (AVVP).
The details of the datasets and the downstream task implementations are provided in the Appendix.
Unless otherwise specified, all methods adopt Swin-V2-Large as the visual backbone and HTS-AT as the audio backbone, following the corresponding baseline configurations to ensure a fair comparison.
% All experiments are conducted on a single NVIDIA RTX 5880 Ada GPU.

\subsection{Implementation Details}

Unless otherwise specified, all experiments were conducted on a single NVIDIA RTX 5880 Ada GPU.
To ensure reproducibility, we maintained consistent data splits and training configurations across tasks, adjusting only the number of experts and loss weights based on task-specific characteristics and convergence stability.

\subsubsection{Number of Experts in AVMoE}

AVMoE freezes large pretrained visual and audio backbones and inserts learnable mixture-of-experts adapters at every layer, including unimodal and cross-modal experts.  Then a lightweight router selects experts based on the spatiotemporal context and aggregates their outputs, preserving backbone stability and parameter efficiency while enabling cross-modal interaction and task adaptation.

Since we freeze the visual and audio backbones, adding unconstrained learnable modules could introduce instability and increase the risk of overfitting during training.
Therefore, in AVMoE, we use a smaller expert scale than the reported optimal scale to balance stability and performance as detailed in Tab.\ref{tab:supp-params}.
It is worth noting that the MUSIC-AVQA dataset contains deliberately introduced cross-modal misalignment samples.
To enhance robustness against such samples, we moderately increase the number of unimodal experts for the AVQA task.

\begin{table}
  \caption{Number of Experts in AVMoE. CMA denotes Cross-Modal Adapters, and UA denotes Unimodal Adapters.}
  \label{tab:supp-params}
  \centering
  \begin{tabular}{ccc}
      \toprule
      Task & Number of CMA & Number of UA  \\
      \midrule
      AVQA & 1             & 2             \\
      AVS  & 1             & 1             \\
      AVE  & 1             & 1             \\
      AVVP & 1             & 1             \\
      \bottomrule
  \end{tabular}
\end{table}

% \begin{table*}
%   \caption{Number of Experts in AVMoE. CMA denotes Cross-Modal Adapters, and UA denotes Unimodal Adapters. At the same time, we recorded the ratio of the total number of parameters to the number of learnable parameters in CAE-AV for the corresponding number of experts.}
%   \label{tab:supp-params}
%   \centering
%   \begin{tabular}{ccccc}
%       \toprule
%       Task & Number of CMA & Number of UA & Trainable Params (\%) & Total Params (M) \\
%       \midrule
%       AVQA & 1             & 2            & 51.7                  & 538.6            \\
%       AVS  & 1             & 1            & 57.7                  & 615.6            \\
%       AVE  & 1             & 1            & 43.5                  & 460.9            \\
%       AVVP & 1             & 1            & 45.2                  & 474.6            \\
%       \bottomrule
%   \end{tabular}
% \end{table*}

\subsubsection{Weight of Loss in CAE-AV}

We define the total loss as the weighted sum of the task's primary loss and three regularization or alignment terms, with weights $\alpha$, $\beta$, and $\gamma$, as shown in Eq.\ref{eq:loss}.
To prevent early instability, $\gamma$ follows a warm-up schedule: it is set to 0 during the initial epochs and then gradually increases to its target value.
The learning rate follows a step decay schedule, decreasing by 35\% every three epochs to balance convergence speed and fine-tuning in later stages.

For the AVQA task, we set $\alpha = 0.08, \beta = 0.03, \gamma = 0.005$, with $\gamma = 0$ for the first five epochs. The initial learning rate is set to $1 \times 10^{-4}$.
For the AVS task, in both settings, we set $\alpha = 0.10, \beta = 0.05, \gamma = 0.005$, with $\gamma = 0$ for the first 15 epochs in the S4 setting and the first 8 epochs in the MS3 setting.
The initial learning rates are set to $3 \times 10^{-4}$ (S4) and $1.5 \times 10^{-4}$ (MS3).
For both AVE and AVVP tasks, we set $\alpha = 0.10, \beta = 0.05, \gamma = 0.003$, with $\gamma = 0$ for the first five epochs.
The initial learning rates are set to $5 \times 10^{-4}$ (AVE) and $3 \times 10^{-4}$ (AVVP).

\begin{table*}
  \caption{\textbf{Audio-Visual Event Localization.} Comparisons on the test set of AVE. The number in \textbf{bold} denotes the best performance.}
  \label{tab:ave}
  \centering
  \small
  \begin{tabular}{cccccc}
      \toprule
      \textbf{Method}      & \textbf{Visual Encoder}                & \textbf{Audio Encoder} & \textbf{Trainable Params(\%) $\downarrow$} & \textbf{Total Params(M) $\downarrow$} & \textbf{Acc}  \\
      \midrule
      PSP~\cite{psp}       & VGG-19                                 & VGG-like               & 0.8                           & 217.4                    & 77.8          \\
      AVEL~\cite{ave}      & ResNet-152                             & VGGish                 & 2.7                              & 136.0                         & 74.0          \\
      CMBS~\cite{cmbs}     & ResNet-152                             & VGGish                 & 6.6                              & 216.7                         & 79.3          \\
      \midrule
      LAVisH~\cite{lavish} & \multicolumn{2}{c}{Swin-V2-B (shared)}                           & 4.4                   & 114.2                              & 78.8                                             \\
      LAVisH~\cite{lavish} & \multicolumn{2}{c}{Swin-V2-L (shared)}                           & 2.7                   & 238.8                              & 81.1                                             \\
      AVMoE~\cite{avmoe} & \multicolumn{2}{c}{Swin-V2-B (shared)}                           & 31.8                   & 150.4                              & 79.4                                             \\
      AVMoE~\cite{avmoe} & \multicolumn{2}{c}{Swin-V2-L (shared)}                           & 32.6                   & 483.1                              & 81.5                                             \\
      \textbf{ours} & \multicolumn{2}{c}{Swin-V2-B (shared)}                           & 45.5                   & 200.7                              & 79.2                                             \\
      \textbf{ours} & \multicolumn{2}{c}{Swin-V2-L (shared)}                           & 35.7                   & 355.9                              & 82.2                                             \\
      \midrule
      LAVisH~\cite{lavish} & Swin-V2-L                              & HTS-AT                 & 30.6                              & 374.9                          & 78.6          \\
      DG-SCT~\cite{dgdct}  & Swin-V2-L                              & HTS-AT                 & 43.6                              & 461.3                         & 82.2          \\
      AVMoE~\cite{avmoe}   & Swin-V2-L                              & HTS-AT                 & 34.9                              & 374.4                         & 82.6          \\
      \midrule
      \textbf{ours}        & Swin-V2-L                              & HTS-AT                 & 43.5                              & 460.9                         & \textbf{82.6} \\
      \bottomrule
  \end{tabular}
\end{table*}

\begin{table*}[t]
  \caption{\textbf{Audio-Visual Video Parsing.} Comparisons on the test set of LLP. The number in \textbf{bold} denotes the best performance and an \underline{underline} denotes the second best result.}
  \label{tab:avvp}
  \centering
  \small
  \begin{tabular}{@{}c*{10}{c}@{}}
      \toprule
      \multirow{2}{*}{\textbf{Method}} & \multicolumn{5}{c}{\textbf{Segment-level}} & \multicolumn{5}{c}{\textbf{Event-level}}                                                                                                                                          \\
      \cmidrule(lr){2-6} \cmidrule(lr){7-11}
                                       & A                                          & V                                        & AV               & Type          & Event         & A             & V                & AV               & Type          & Event         \\
      \midrule
      AVEL~\cite{ave}                  & 49.9                                       & 37.3                                     & 37.0             & 41.4          & 43.6          & 43.6          & 32.4             & 32.6             & 36.2          & 37.4          \\
      AVSDN~\cite{avsdn}               & 47.8                                       & 52.0                                     & 37.1             & 45.7          & 50.8          & 34.1          & 46.3             & 26.5             & 35.6          & 37.7          \\
      HAN~\cite{han}                   & 60.1                                       & 52.9                                     & 48.9             & 54.0          & 55.4          & 51.3          & 48.9             & 43.0             & 47.7          & 48.0          \\
      MGN~\cite{mgn}                   & 60.7                                       & 55.5                                     & 50.6             & 55.6          & 57.2          & 51.0          & 52.4             & 44.4             & 49.3          & 49.2          \\
      DG-SCT~\cite{dgdct}              & 59.0                                       & 59.4                                     & 52.8             & 57.1          & 57.0          & 49.2          & \textbf{56.1}    & 46.1             & 50.5          & 49.1          \\
      AVMoE~\cite{avmoe}               & 62.1                                       & \textbf{60.0}                            & \textbf{54.4}    & 58.8          & 59.0          & 51.8          & 55.7             & \textbf{47.6}    & 51.7          & 50.2          \\
      \midrule
      \textbf{ours}                    & \textbf{63.8}                              & \underline{59.6}                         & \underline{53.9} & \textbf{59.0} & \textbf{60.4} & \textbf{53.3} & \underline{56.0} & \underline{47.0} & \textbf{52.8} & \textbf{51.6} \\
      \bottomrule
  \end{tabular}
\end{table*}

\begin{table}
  \caption{\textbf{Audio-Visual Question Answering.} Comparisons on the test set of MUSIC-AVQA. The number in \textbf{bold} denotes the best performance and an \underline{underline} denotes the second best result.}
  \label{tab:avqa}
  \centering
  \small
  \begin{tabular}{@{}ccccc@{}}
      \toprule
      \textbf{Method}                & \textbf{AQ}   & \textbf{VQ}   & \textbf{AVQ}     & \textbf{Avg}  \\
      \midrule
      AVSD~\cite{schwartz2019simple} & 68.5          & 70.8          & 65.5             & 67.4          \\
      Pano-AVQA~\cite{pano}          & 70.7          & 72.6          & 66.6             & 68.9          \\
      ST-AVQA~\cite{musicavqa}       & 74.1          & 74.0          & 69.5             & 71.5          \\
      LAVisH~\cite{lavish}           & 75.7          & 80.4          & 70.4             & 74.0          \\
      LAVisH~\cite{lavish}           & 75.4          & 79.6          & 70.1             & 73.6          \\
      DG-SCT~\cite{dgdct}            & 77.4          & 81.9          & 70.7             & 74.8          \\
      CoPL~\cite{copl}               & 76.8          & 77.1          & \textbf{75.2}    & 75.9          \\
      AVMoE~\cite{avmoe}             & 77.6          & 82.7          & 71.9             & 75.7          \\
      \midrule
      \textbf{ours}                  & \textbf{78.2} & \textbf{83.9} & \underline{72.7} & \textbf{76.7} \\
      \bottomrule
  \end{tabular}
\end{table}

\begin{table}
  \caption{\textbf{Audio-Visual Segmentation.} Comparisons on the S4 and MS3 settings of AVSBench. The number in \textbf{bold} denotes the best performance.}
  \label{tab:avs}
  \centering
  \small
  \begin{tabular}{c c c c c}
      \toprule
      \multirow{2}{*}{\textbf{Model}} & \multicolumn{2}{c}{\textbf{S4}} & \multicolumn{2}{c}{\textbf{MS3}}                                     \\
      \cmidrule(lr){2-3} \cmidrule(lr){4-5}
                                      & $\mathcal{M_J}$                 & $\mathcal{M_F}$                  & $\mathcal{M_J}$ & $\mathcal{M_F}$ \\
      \midrule
      AVS~\cite{avs}                  & 78.7                            & 87.9                             & 54.0            & 64.5            \\
      LAVisH~\cite{lavish}            & 80.1                            & 88.0                             & 49.8            & 60.3            \\
      LAVisH~\cite{lavish}            & 78.0                            & 87.0                             & 49.1            & 59.9            \\
      CoPL~\cite{copl}                & 80.1                            & -                                & -               & -               \\
      DG-SCT~\cite{dgdct}             & 80.9                            & 89.2                             & 53.5            & 64.2            \\
      AVMoE~\cite{avmoe}              & 81.1                            & 89.7                             & 54.5            & 68.7            \\
      \midrule
      \textbf{ours}                   & \textbf{81.9}                   & \textbf{89.9}                    & \textbf{55.1}   & \textbf{69.2}   \\
      \bottomrule
  \end{tabular}
\end{table}

% \usepackage{multirow}
\begin{table*}
  \caption{CAE-AV ablation experiments. We selectively present the most representative metrics from four downstream tasks.}
  \label{tab:ablation}
  \centering
  \small
  \begin{tabular}{cccccccccccccc}
      \toprule
      \multicolumn{3}{c}{\textbf{Module}} & \multicolumn{4}{c}{\textbf{AVQA}} & \multicolumn{4}{c}{\textbf{AVS}} & \textbf{AVE}        & \multicolumn{2}{c}{\textbf{AVVP}}                                                                                                                                                                                                             \\
      \cmidrule(lr){1-3} \cmidrule(lr){4-14}
      \multirow{2}{*}{CASTE}              & \multirow{2}{*}{CASE}             & \multirow{2}{*}{Loss}            & \multirow{2}{*}{AQ} & \multirow{2}{*}{VA}               & \multirow{2}{*}{AVQ} & \multirow{2}{*}{Avg} & \multicolumn{2}{c}{S4} & \multicolumn{2}{c}{ms3} & \multirow{2}{*}{Acc} & \multicolumn{2}{c}{Segment-level}                                                 \\
                                          &                                   &                                  &                     &                                   &                      &                      & $\mathcal{M_J}$        & $\mathcal{M_F}$         & $\mathcal{M_J}$      & $\mathcal{M_F}$                   &               & Type          & Event         \\
      \midrule
      -                                   & -                                 & -                                & 77.5                & 81.3                              & 70.9                 & 74.8                 & 80.8                   & 89.5                    & 54.3                 & 68.2                              & 79.4          & 55.9          & 57.2          \\
      \checkmark                          & -                                 & -                                & 76.4                & 83.7                              & 72.3                 & 76.0                 & 81.0                   & 89.7                    & 55.0                 & 69.1                              & 80.2          & 58.3          & 59.5          \\
      -                                   & \checkmark                        & -                                & 77.8                & 83.5                              & 72.6                 & 76.4                 & 81.8                   & 89.9                    & 54.4                 & 68.5                              & 82.2          & 56.8          & 59.3          \\
      \checkmark                          & \checkmark                        & -                                & 78.0                & 83.8                              & 72.6                 & 76.6                 & 81.7                   & 89.7                    & 55.0                 & 69.2                              & 82.6          & 58.5          & 60.3          \\
      \midrule
      \checkmark                          & \checkmark                        & \checkmark                       & \textbf{78.2}       & \textbf{83.9}                     & \textbf{72.7}        & \textbf{76.7}        & \textbf{81.9}          & \textbf{89.9}           & \textbf{55.1}        & \textbf{69.2}                     & \textbf{82.6} & \textbf{59.0} & \textbf{60.4} \\
      \bottomrule
  \end{tabular}
\end{table*}

\subsection{Comparison Experiments}

We evaluate CAE-AV on the AVE benchmark~\cite{ave}, Tab.\ref{tab:ave} shows that CAE-AV achieves an accuracy of 82.6\% under the same Swin-V2-L and HTS-AT settings as AVMoE, matching the current state-of-the-art frozen backbone method ~\cite{avmoe}.

On the LLP dataset~\cite{avvp}, we report multiple metrics at both segment-level and event-level, with results shown in Tab.\ref{tab:avvp}. CAE-AV surpassed the current SOTA AVMoE by 1.7\% in the audio type at the Segment-level, and by 0.2\% and 0.6\% in the Type and event. At the Event-level, the audio, type, and event exceeded AVMoE by 1.5\%, 1.1\%, and 1.4\%, respectively. Although a more conservative injection strategy on the most stringent AV channel resulted in a slight decline in metrics, the overall improvement in Type and Event consistency demonstrates that CAE-AV excels at making stable and interpretable decisions in weakly supervised parsing.

On the MUSIC-AVQA dataset~\cite{musicavqa}, we evaluate CAE-AV across three question types: audio-only (AQ), visual-only (VQ), and audio-visual (AVQ). As shown in Tab.\ref{tab:avqa}, CAE-AV achieves an average accuracy of 76.7%, surpassing the current SOTA AVMoE by 1.0%. Consistent improvements across all three question types further verify the effectiveness of our approach. By selectively injecting informative cues and leveraging caption-aligned semantic guidance, CASTE and CASE jointly enhance key-frame selection and semantic generalization in multimodal question answering.

On AVSBench~\cite{avs}, we evaluate CAE-AV under the single-source (S4) and multi-source (MS3) settings using region similarity $\mathcal{J}$ (Jaccard index) and contour accuracy $\mathcal{F}$ (F-score). As shown in Tab.\ref{tab:avs}, CAE-AV consistently outperforms current SOTA AVMoE across both settings, achieving gains of +0.8 $\mathcal{M_J}$ / +0.2 $\mathcal{M_F}$ on S4 and +0.6 $\mathcal{M_J}$ / +0.5 $\mathcal{M_F}$ on MS3. In multi-source scenarios, the frame-level consistency gate in CASTE effectively mitigates misalignment caused by off-screen sources and shot transitions. Meanwhile, CASE provides semantic guidance that enables the model to maintain stable focus on sounding regions under occlusion and reverberation.

% , particularly within multi-event and noisy environments
\subsection{Ablation Study}

% We conduct extensive ablation studies to verify the effectiveness of CASTE, CASE, and the proposed loss design.
% As shown in \ref{tab:ablation}, adding either CASTE or CASE individually to the baseline yields substantial gains across tasks, confirming that each module contributes independently.
% Combining CASTE and CASE produces further improvements on most metrics, indicating complementary effects: agreement-guided spatiotemporal enrichment reduces cross-modal misalignment, while caption-aligned injection sharpens semantic focus.
% Finally, introducing the four loss functions on top of CASE controls the strength and sparsity of caption injection, stabilizes optimization, and delivers the largest overall gains.

We conducted extensive ablation studies to verify the effectiveness of CASTE, CASE, and the proposed loss design.
The first line is the result of the AVMoE\cite{avmoe} we reproduced, please refer to Supplementary Section 3.1 for specific details.
As shown in Tab.\ref{tab:ablation}, adding CASTE to the baseline yields clear gains, including a 2.4\% increase in AVVP segment-level Type and a 2.3\% improvement in Event accuracy, along with consistent enhancements in AVS and AVE accuracy.
Adding CASE alone yields greater improvements on alignment-heavy metrics, including a 1.6\% increase in AVQA average accuracy, a 1.0\% increase in AVS S4 $\mathcal{M_J}$, and a 2.8\% increase in AVE accuracy.
Combining CASTE and CASE further amplifies these trends, resulting in a 3.2\% increase in AVE accuracy and improvements of 3.1\% and 2.6\% in AVVP segment-level Event and Type, respectively. This indicates complementary effects of CASTE and CASE, where agreement-guided spatiotemporal enrichment reduces cross-modal misalignment, while caption-aligned injection sharpens semantic focus.
Introducing four loss functions on top of the combined model stabilizes the gains and continues to improve several metrics. For example, there is a 0.5\% increase in AVVP segment-level Type and a 0.2\% increase in AVS S4 $\mathcal{M_J}$, all while maintaining the best overall results.

\begin{figure}
  \centering
  \includegraphics[width=1.0\linewidth]{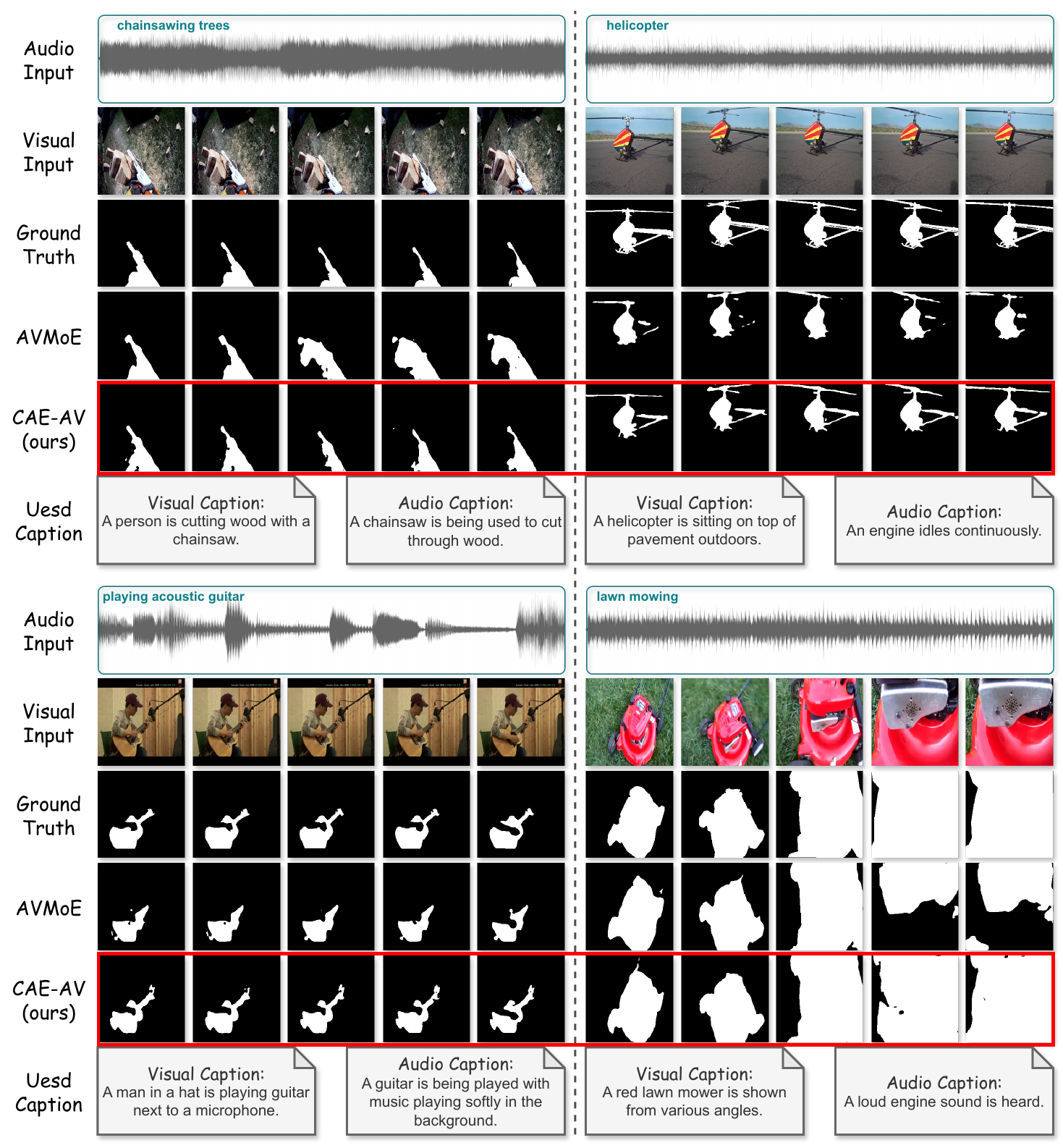}
  \caption{Qualiative examples of the AVMoE and our CAE-AV, under the S4 setting of the AVS task. The first and second rows of each subgraph represent the Audio Input and Visual Input, respectively. The third row shows the Ground Truth, while the fourth and fifth rows display the results of the AVMoE and our CAE-AV in locating and outlining object shapes. The sixth row presents the visual caption and audio caption used in CAE-AV to provide semantic information to the model.}
  \label{fig:supp-s4}
\end{figure}

\subsection{Qualitative Experiments of Audio-Visual Segmentation}

We provide qualitative examples from the AVS task to evaluate the effectiveness of the CAE-AV model.
As shown in Fig.\ref{fig:supp-s4}, under the S4 setting, in the ``chainsawing trees'' scenario, the AVMoE method exhibits significant localization errors in frames 3, 4, and 5, where the segmentation regions extend beyond the actual target, mistakenly including the surrounding trees.
The semantic information in the Visual Caption, ``wood with a chainsaw,'' aids our CAE-AV in accurately distinguishing between the chainsaw and the wood, avoiding such missegmentation.
In the ``helicopter'' scenario, AVMoE struggles with segmentation completeness, particularly failing to separate the helicopter's body and rotor clearly.
In contrast, CAE-AV produces a clean segmentation with clear contours for both the body and rotor.
In the ``playing acoustic guitar'' scenario, AVMoE's segmentation misses key details of the guitar, particularly in areas like the neck, and underrepresents the overall size of the guitar.
Our CAE-AV, however, successfully segments the entire structure of the guitar.
These examples demonstrate that the spatial enrichment in the CASTE module helps the model focus on detailed regions.
In the ``lawn mowing'' scenario, after a viewpoint change in frame 4, AVMoE's segmentation completely deviates from the target lawnmower.
However, with the guidance from the Visual Caption's phrases, ``a red lawn mower'' and ``various angles,'' our CAE-AV accurately identifies the lawnmower even under the altered viewpoint, highlighting the positive impact of the CASE module's high-level semantic guidance in complex scenarios.

\begin{figure}
  \centering
  \includegraphics[width=1.0\linewidth]{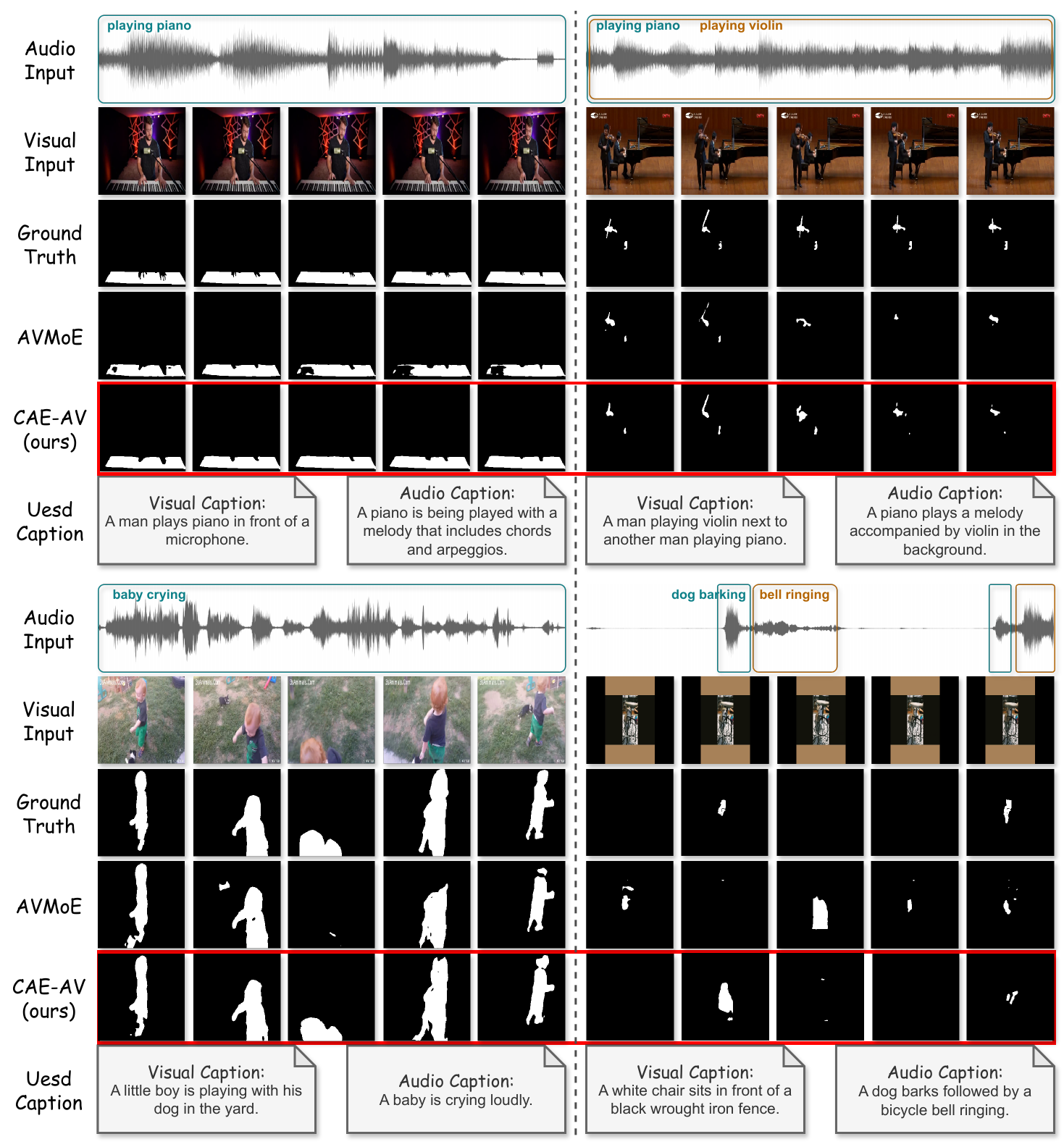}
  \caption{Qualiative examples of the AVMoE and our CAE-AV, under the MS3 setting of the AVS task.}
  \label{fig:supp-ms3}
\end{figure}

As shown in Fig.\ref{fig:supp-ms3}, under the MS3 setting, in the ``playing piano'' scenario, AVMoE fails to capture key details of the piano keys in frames 1, 3, 4, and 5, whereas the piano region segmented by CAE-AV closely matches the ground truth.
In the ``playing piano and playing violin'' scene, AVMoE fails to outline the shape of the piano in frames 3 and 4, while the violin region progressively shifts away from the actual object.
Our CAE-AV, on the other hand, produces a much clearer and more accurate violin contour, and the semantic information from the Visual Caption, ``A man... next to another man...,'' helps CAE-AV differentiate between the two targets, ``piano'' and ``violin.''
Thanks to CASTE's modeling of temporal frame information, CAE-AV maintains partial segmentation despite occlusion of the piano in the later frames, inferring results from earlier frames.
In the ``baby crying'' scenario, AVMoE incorrectly includes the dog in the segmentation area in frames 1 and 2, and completely loses the baby target in frame 3.
The Audio Caption's mention of ``baby'' and absence of ``dog'' helps CAE-AV consistently and precisely locate the baby.
Furthermore, when the baby moves out of the video frame in the third frame, CAE-AV is still able to infer the baby's location by integrating information from surrounding frames.
In the ``dog barking and bell ringing'' scenario, AVMoE produces significant false-positive segments in frames 1, 3, and 4, despite no sound or sound sources being present in the visual scene.
Although CAE-AV also makes an error in frame 2, where it mistakenly merges the dog and chair regions, we attribute this missegmentation to insufficient semantic information from the caption, which did not fully capture the distinctive features of the ``dog.''
In future work, we plan to explore using longer, more detailed descriptions to improve caption-based guidance for better segmentation accuracy.

\subsection{Qualitative Experiments of Audio-Visual Video Parsing}

\begin{figure}
  \centering
  \includegraphics[width=1.0\linewidth]{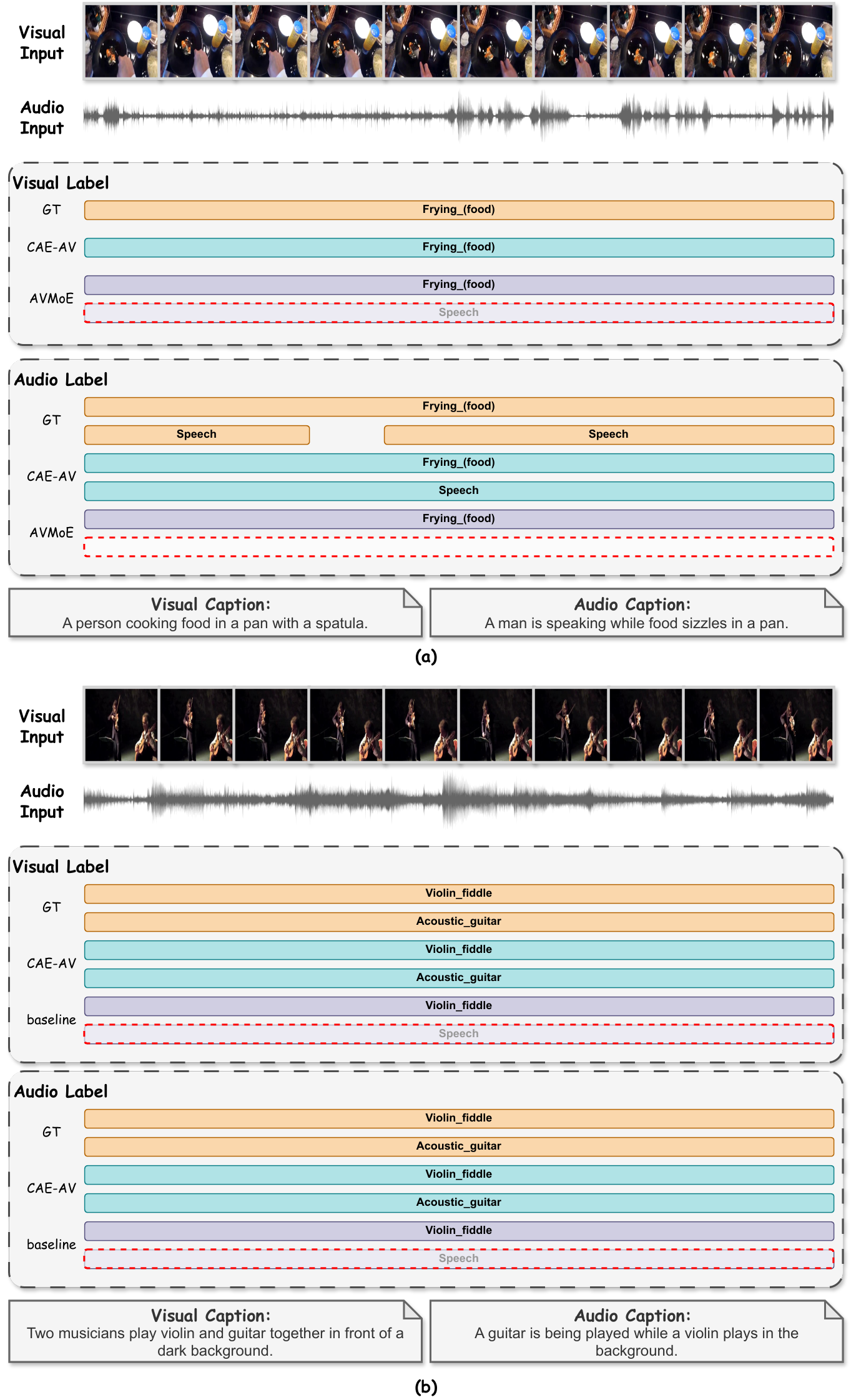}
  \caption{Qualiative examples of the AVMoE and our CAE-AV, under the AVVP task. The first and second rows in each subgraph represent visual and audio inputs, respectively. Each row in the corresponding modal label represents the ground truth, the results of our CAE-AV and AVMOE. Finally, the visual caption and audio caption used in the corresponding examples were presented.}
  \label{fig:supp-avvp}
\end{figure}

We provide some qualitative examples from the AVVP task to evaluate the effectiveness of the CAE-AV model.
As shown in Fig.\ref{fig:supp-avvp}(a), the visual input clearly depicts ``frying food,'' but AVMoE erroneously associates the hand with ``speech,'' resulting in the incorrect appearance of the ``speech'' label in the visual segmentation.
However, the ``speech'' label is missing in the audio segmentation.
In contrast, our CAE-AV benefits from the caption, where the visual caption does not mention anything related to ``speech,'' but the audio caption explicitly includes the term ``speaking.''
This demonstrates that the advanced semantic guidance in CASE successfully influences feature distribution.
Similarly, in Fig.\ref{fig:supp-avvp}(b), when the guitar is not the main visual focus, AVMoE fails to detect the ``guitar'' label, while CAE-AV correctly identifies the guitar despite its subtle presence in the visual scene.

\section{Conclusion}
\label{sec:conclusion}

In this work, we present CAE-AV, a novel framework designed to address audio-visual misalignment in AVL. Built upon frozen backbones, CAE-AV achieves selective cross-modal enhancement through two lightweight yet complementary modules. The CASTE module adaptively balances spatial and temporal enrichment via agreement consistency and applies residual modulation only to salient tokens, effectively suppressing noise propagation. The CASE module leverages CLIP-encoded visual and audio captions as semantic priors to guide cross-attention and lightweight temporal consolidation. Comprehensive evaluations on AVE, AVVP, AVS, and AVQA validate the effectiveness of CAE-AV, which delivers state-of-the-art performance across diverse tasks while maintaining remarkable parameter efficiency and stable optimization. 
In the future, we will continue to address issues of multimodal misalignment in open-world audio-visual scenarios.

\bibliographystyle{IEEEtran}
\bibliography{main}

\clearpage

\appendices

\section{Generation of Caption}
\begin{figure}[!h]
  \centering
  \includegraphics[width=1.0\linewidth]{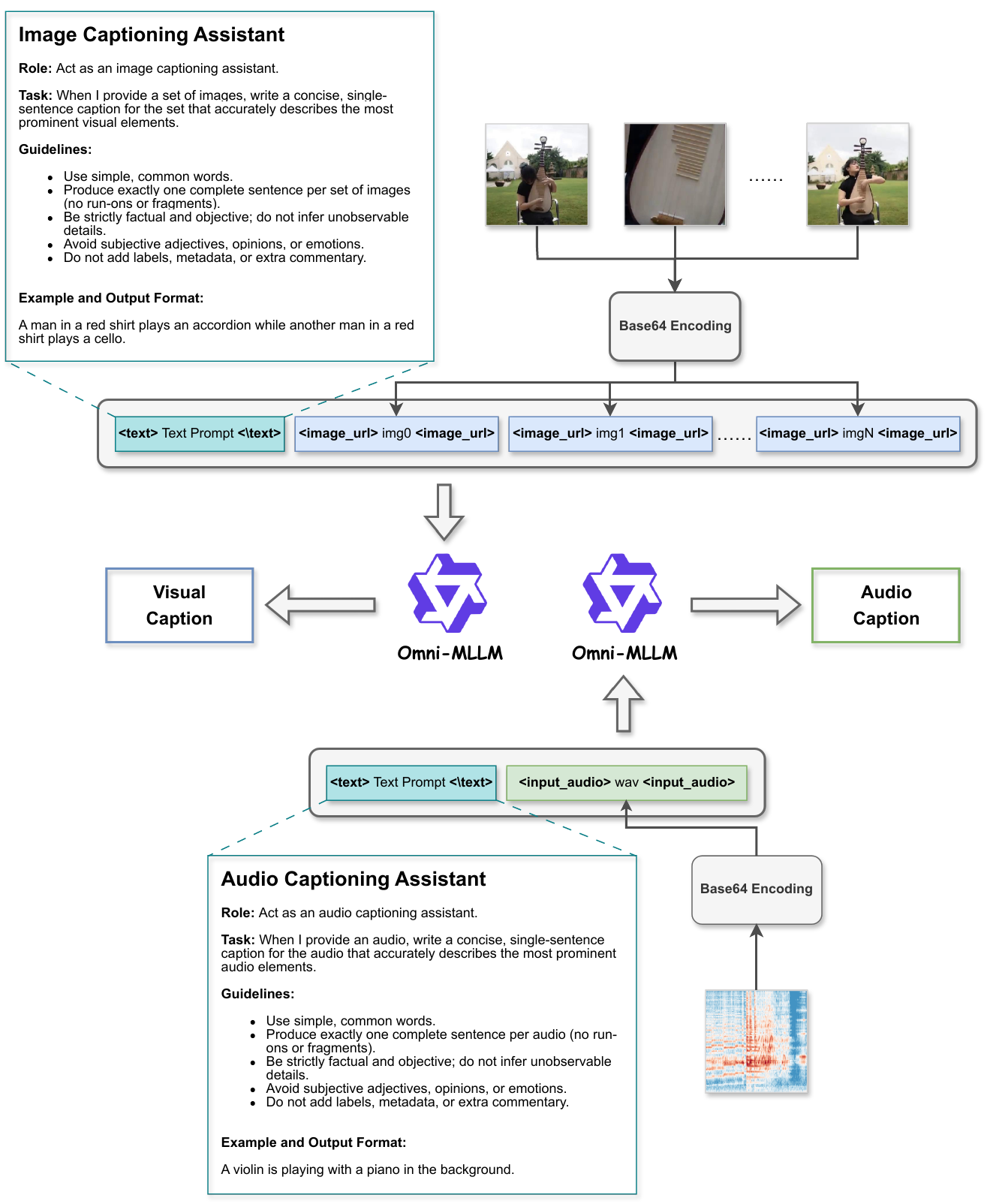}
  \caption{The process of CAE-AV obtaining visual and audio captions.}
  \label{fig:supp-caption}
\end{figure}

As shown in \ref{fig:supp-caption}, our caption generation process consists of two branches.
To ensure consistency and comparability of the prompts, we pre-construct two structurally symmetrical Markdown texts as prompt templates for the Qwen-2.5-Omni multimodal large language model (MLLM)~\cite{qwen}.
These templates, illustrated in the figure, are designed with symmetrical content and format, differing only in modality-specific wording to minimize prompt bias and enhance experimental reproducibility.

For the visual branch, we sample video frames according to the sampling method specific to each downstream task. Each frame is base64-encoded and inserted into a sequence of $\langle$\textit{image\_url}$\rangle$ tags in temporal order. This sequence is then concatenated with the corresponding text template to form the complete input, which is fed into the MLLM to generate the corresponding visual caption.
For the audio branch, we base64-encode the entire audio segment and embed it within an $\langle$\textit{input\_audio}$\rangle$ tag. This is then concatenated with the corresponding text template before being input into the MLLM to generate the audio caption.

\begin{table*}
  \caption{Dataset statistics. Each dataset displays the number of videos and annotated frames.}
  \label{tab:supp-dataset}
  \centering
  \begin{tabular}{c c c c c c}
      \toprule
      \textbf{Datasets}           & \textbf{Videos} & \textbf{Frames} & \textbf{Classes/Answers} & \textbf{Types} & \textbf{Annotations} \\
      \midrule
      MUSIC-AVQA~\cite{musicavqa} & 9,288           & 45,867          & 42                       & video          & answer               \\
      AVSBench~\cite{avs}         & 5,356           & 12,972          & 23                       & video          & pixel                \\
      AVE~\cite{ave}              & 4,143           & 41,430          & 28                       & video          & category             \\
      LLP~\cite{avvp}             & 11,849          & 11,849          & 25                       & video          & category             \\
      \bottomrule
  \end{tabular}
\end{table*}

\section{Downstream Tasks and Datasets}
\begin{figure}[!h]
  \centering
  \includegraphics[width=1.0\linewidth]{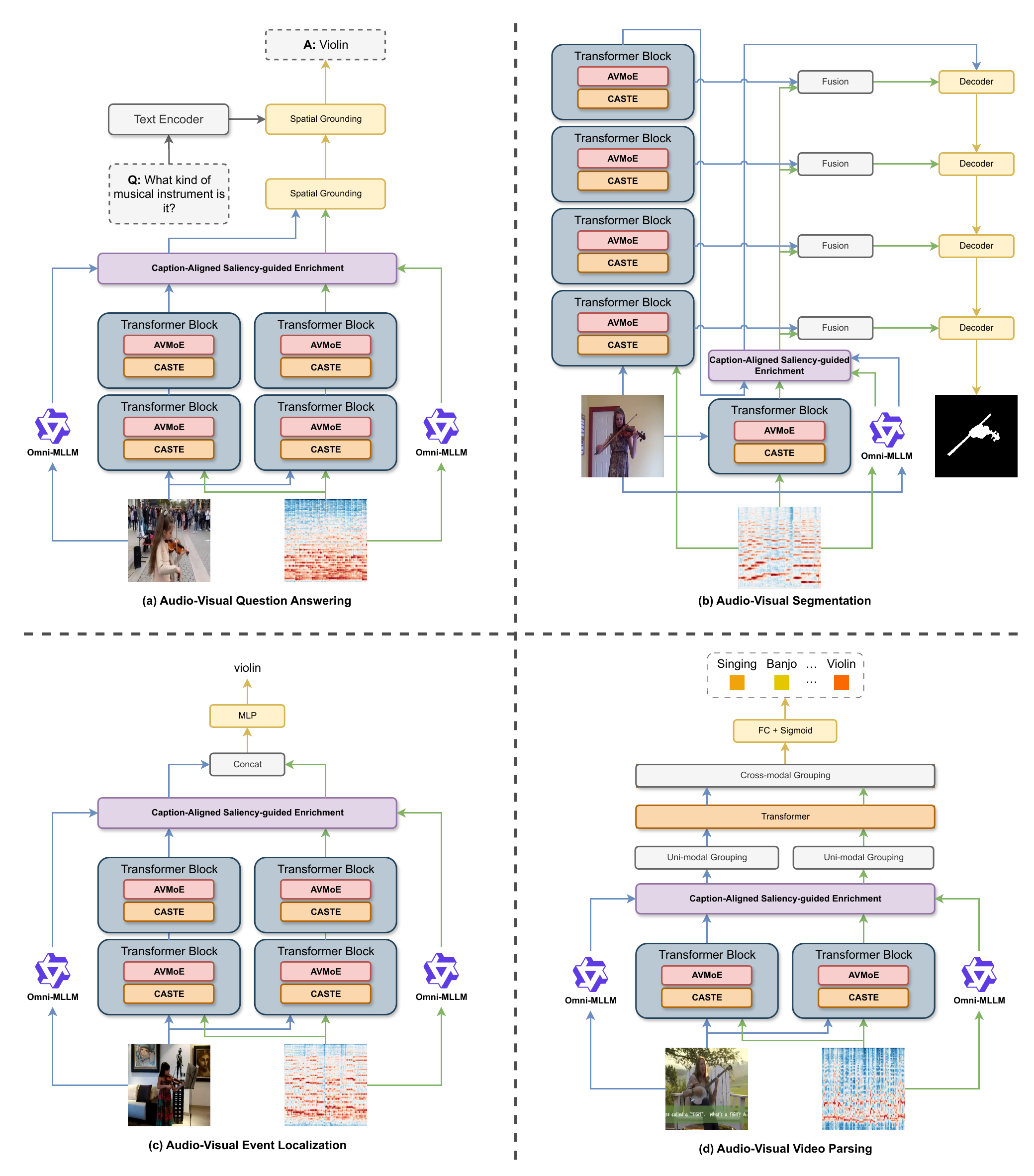}
  \caption{CAE-AV framework diagram adapted for four distinct downstream tasks.}
  \label{fig:supp-down}
\end{figure}

To verify the applicability and robustness of the proposed CAE-AV across multiple types of audio-visual tasks, we conducted experiments on four representative downstream tasks:
audio-visual question answering (AVQA), audio-visual segmentation (AVS), audio-visual event localization (AVE) and audio-visual video parsing (AVVP).
The overall adaptation framework is shown in \ref{fig:supp-down}. Unless otherwise specified, the visual backbone is Swin-V2-L, the audio backbone is HTS-AT, the backbone remains frozen, and only parameter efficient modules are introduced to ensure comparability.
The statistics of each dataset are shown in \ref{tab:supp-dataset}.

\subsection{Audio-Visual Question Answering}

AVQA aims to answer natural language questions about visible objects, sound sources, and their relationships within a video.
We conduct experiments on the MUSIC-AVQA dataset~\cite{musicavqa}, which contains over 45,000 question-answer pairs spanning 33 question templates and features complex scenes and sound source distributions.
For this task, we insert trainable modules into frozen visual and audio Transformer, replacing the pretrained encoders used in the baseline.
This setup enables features and questions to interact within a unified space, followed by a classification head that generates answers, as illustrated in \ref{fig:supp-down}(a).
We use answer prediction accuracy as the evaluation metric, consistent with existing settings~\cite{musicavqa}.

\subsection{Audio-Visual Segmentation}

AVS focuses on pixel-level segmentation of audio-driven sound sources.
Experiments are conducted on AVSBench~\cite{avs},
which includes two settings: single-source (S4) and multi-source (MS3).
The S4 setting consists of 4,932 semi-supervised videos, where only the first frame has pixel-level masks, but evaluation requires predictions for all frames.
The MS3 setting includes 424 fully supervised videos, with annotations for every frame.
Following the pipeline in \ref{fig:supp-down}(b), we connect the frozen visual and audio backbones with trainable modules that replace the original encoding branches.
The fused features are then passed into a decoder to generate the segmentation results.
Evaluation metrics include region similarity $\mathcal{J}$ (Jaccard index) and contour accuracy $\mathcal{F}$ (F-score), consistent with prior work~\cite{avs}.

\subsection{Audio-Visual Event Localization}

AVE focuses on events that are simultaneously visible and audible along the time axis~\cite{ave}.
The dataset includes 4,143 videos categorized into 28 classes, with each video lasting no longer than 10 seconds and containing at least one 2-second audio-visual event.
Each video is uniformly divided into 10 consecutive time segments, from which audio and visual features are extracted using parameter-efficient modules integrated into frozen backbones.
Cross-modal interaction is explicitly modeled, and the concatenated audio-visual representations are fed into a two-layer MLP for segment-level classification, as illustrated in \ref{fig:supp-down}(c).
Following standard practice~\cite{cmbs}, we use the proportion of correctly localized time segments (accuracy) as the evaluation metric.

\subsection{Audio-Visual Video Parsing}

AVVP aims to segment a video into a series of temporal intervals and classify each segment as audible (A), visible (V), or audio-visual (AV).
We use the LLP dataset~\cite{avvp}, which contains 11,849 videos from diverse domains, categorized into 25 classes.
The dataset provides video-level event annotations, with 1,849 segments further annotated at the per-second level, creating a semi-supervised learning setting.
For implementation, we build upon MGN~\cite{mgn} and introduce parameter-efficient modules within its cross-modal fusion Transformer layers to enhance representation learning, as illustrated in \ref{fig:supp-down}(d).
Following existing evaluation practices~\cite{mgn}, we report segment-level and event-level A/V/AV metrics, as well as the overall Type and Event F-scores.

\section{Ablation experiments at different locations of CASTE}

To investigate the impact of different insertion positions of CAST on the network's final performance, we conducted corresponding ablation experiments. As shown in \ref{tab:castelocation}, Location 1 represents placing CASTE before two AVMoE modules; Location 2 represents inserting CASTE between the two modules; and Location 3 represents placing CASTE after the AVMoE modules. The experimental results indicate that placing CASTE before AVMoE achieves better performance. This is because CAST, as a front-end module, can conditionally align multimodal features earlier, highlight task-relevant cues, suppress irrelevant noise, and provide more discriminative inputs for subsequent AVMoE routing and expert activation. When CAST is located in the middle or at the end, AVMoE has already completed information aggregation and expert allocation on features that are not fully aligned, leaving limited room for subsequent correction and resulting in weaker overall performance.

\begin{table}
    \caption{Ablation experiments on the performance of CASTE at different positions for both the AVQA and AVE tasks. The number in \textbf{bold} denotes the best performance.}
    \label{tab:castelocation}
    \centering
    \begin{tabular}{ccc}
        \toprule
        Locations & AVQA(Avg)     & AVE           \\
        \midrule
        \textbf{Location 1 (CAE-AV)}    & \textbf{76.7} & \textbf{82.6} \\
        Location 2    & 76.5          & 80.5          \\
        Location 3    & 75.8          & 81.4          \\
        \bottomrule
    \end{tabular}
\end{table}

% \section*{Acknowledgment}


\ifCLASSOPTIONcaptionsoff
  \newpage
\fi

\end{document}